\documentclass{article} 
\usepackage{iclr2016_workshop,times}
\usepackage{hyperref}
\usepackage{url}
\usepackage{graphicx}
\usepackage{subfig}
\usepackage{natbib}

\title{Learning Representations of Affect from Speech }

\author{Sayan Ghosh, Eugene Laksana\\
Institute for Creative Technologies \\
Department of Computer Science\\
University of Southern California \\
Los Angeles, CA, USA \\
\texttt{\{sghosh,elaksana\}@ict.usc.edu} \\
\And
Louis-Philippe Morency \\
Language Technologies Institute \\
Carnegie-Mellon University \\
Pittsburgh, PA, USA \\
\texttt{morency@cs.cmu.edu} \\
\And
Stefan Scherer\\
Institute for Creative Technologies \\
Department of Computer Science\\
University of Southern California \\
Los Angeles, CA, USA \\
\texttt{scherer@ict.usc.edu} \\
}

\begin{document}

\maketitle

\begin{abstract}
There has been a lot of prior work on representation learning for speech recognition applications, but not much emphasis has been given to an investigation of effective representations of affect from speech, where the paralinguistic elements of speech are separated out from the verbal content. In this paper, we explore denoising autoencoders for learning paralinguistic attributes, i.e. categorical and dimensional affective traits from speech. We show that the representations learnt by the bottleneck layer of the autoencoder are highly discriminative of activation intensity and at separating out negative valence (sadness and anger) from positive valence (happiness). We experiment with different input speech features (such as FFT and log-mel spectrograms with temporal context windows), and different autoencoder architectures (such as stacked and deep autoencoders). We also learn utterance specific representations by a combination of denoising autoencoders and BLSTM based recurrent autoencoders. Emotion classification is performed with the learnt temporal/dynamic representations to evaluate the quality of the representations. Experiments on a well-established real-life speech dataset (IEMOCAP) show that the learnt representations are comparable to state of the art feature extractors (such as voice quality features and MFCCs) and are competitive with state-of-the-art approaches at emotion and dimensional affect recognition.
\end{abstract}

\section{Introduction}

The field of representation learning has witnessed a huge progress over recent years, particularly in the automated machine perception of images, text and speech. This has been facilitated by developments in deep learning, where it has been observed that stacking layers in neural networks with huge amounts of data has yielded classification accuracies which have beaten the state-of-the-art in large scale recognition problems, such as ILSVRC challenge~\citep{russakovsky2014imagenet}, and large scale speech recognition~\citep{hinton2012deep}. However, there has been limited progress in exploring representation learning for non-phonetic and affective attributes from speech. Current research is mainly on the use of off-the-self feature extractors, such as Mel-Frequency Cepstral Coefficients (MFCCs), and voice quality features~\citep{scherer2013investigating} such as normalized amplitude quotient, and fundamental frequency. With this view in mind, we wish to explore the effectiveness of representation learning for affect from speech. We select not only utterance-level categorical emotions, but also activation and valence as attributes of interest. Activation is a measure of a subject's excitability when communicating emotion. High activation would correspond to emotions such as excitement and anger, while low activation would be displayed in sadness. Valence is the measure of the subject's sentiment, where anger and sadness correspond to negative valence, while happiness corresponds to positive valence. \\

We primarily focus on two models - (1) Denoising autoencoder learnt from the speech spectrogram. We employ both log-frequency and log-mel spectrograms to learn representations, and also investigate the effect of windowed frames to encode contextual information. This is motivated by prior work, which compute frame-level feature statistics over long temporal sliding windows~\citep{kim2013emotion}. Both stacked (with tied/untied weights) and deep autoencoder architectures are trained, and the best spectrogram-autoencoder configuration is identified by classification experiments on a speaker independent hold-out validation set. (2) Recurrent autoencoder, where the entire feature sequence for each utterance is reconstructed by a recurrent neural network. We further train a BLSTM-RNN as an autoencoder over the activations obtained for selected spectrogram and autoencoder configurations, and show that the utterance level representations from speech effectively capture paralinguistic attributes in the data.

\section{Prior Work in Representation Learning from Speech}

~\citet{jaitly2011new} introduced transforming autoencoders to learn acoustic events in terms of onset times, amplitudes, and rates. They use the Arctic database and TIMIT for their experiments. ~\citet{graves2013speech} investigated deep recurrent neural networks for speech recognition, achieving a test set error of 17.7\% on TIMIT. ~\citet{sainath2013learning} worked on filterbank learning in a deep neural network framework, where it was found that filterbanks similar to the Mel scale were learnt. They improved on their work in ~\citet{sainath2014improvements} by incorporating delta learning and speaker adaptation. ~\citet{han2014speech} performed speech emotion recognition from the IEMOCAP corpus using a combination of DNN (Deep Neural Network) and Extreme Learning Machines. They obtained 20\% relative accuracy improvement compared to state-of-the-art approaches. However the prior work on speech emotion recognition using neural networks has mostly focused on the usage of standard feature extractors, such as MFCCs, pitch and zero-crossing rate, and how these features correlate with emotion and affective attributes~\citep{lee2011emotion}. There has been very little work done on representation learning on lower level features, such as the time domain waveform/spectrogram for this application. To the best of our knowledge, our work is the first attempt to investigate emotional attributes from an unsupervised analysis of spectrograms. 

\section{Models and Experimental Setup}
\subsection{Denoising Autoencoder}

The autoencoder~\citep{baldi2012autoencoders} is a neural network typically trained to learn a lower-dimensional distributed representation of the input data. The input dataset of $N$ data points $\{\mathbf{x}_i\}_{i=1}^{i=N}$ is passed into a feedforward neural network of one hidden layer, where the hidden layer is a bottleneck layer with activations $\{y_i\}_{i=1}^{i=N}$. The activations are obtained as (for the purposes of this paper, we assume $tanh$ activations):
	\[\mathbf{y}_i = \tanh{(\mathbf{Wx}_i+\mathbf{b})}
\]
The output of the autoencoder $\mathbf{z}_i$ is obtained from the autoencoder activations as:
	\[\mathbf{z}_i = \mathbf{W'y}_i+\mathbf{b'}
\] 
For an autoencoder with tied weights, we have $\mathbf{W} = \mathbf{W'}^{T}$. This would require less parameters to be trained, thus acting as a regularizer. In this paper, we investigate the quality of representations using both tied and untied weights, since it is not clearly evident which configuration is better. Thus the autoencoder is trained using backpropagation, much as in an ordinary feedforward neural network. In this paper, the loss function employed for training the autoencoder is the SSE (Sum of Squared Error Loss) $L=\sum_{i=1}^{i=N} \|\mathbf{x}_i-\mathbf{z}_i\|^2$. ~\citet{vincent2008extracting} introduce denoising autoencoders, where the data point $\mathbf{{x}}_i$ is corrupted (by randomly setting a fraction of the elements to zero) to produce $\mathbf{\tilde{x}}_i$, from which the original clean data point $\mathbf{{x}}_i$ is reconstructed by the autoencoder.  This enables the autoencoder to learn not only latent information in the data, but also robust dependencies among the elements in $\mathbf{{x}}_i$. 
When training denoising autoencoders in a greedy stacked fashion, we have multiple layers with weights $\mathbf{W}^{k-1}$ and $\mathbf{W}^{k}$ for the $k$-th hidden layer, where the autoencoder activations at the layer are (where we have $\mathbf{y}_{i}^{0}=\mathbf{\tilde{x}}_i$) :
\[
\mathbf{y}_{i}^{k} = \tanh{(\mathbf{W}^{k-1}\mathbf{y}_{i}^{k-1}+\mathbf{b}^{k-1})}
\]

\subsection{Deep Autoencoder}

In a deep autoencoder, the network is trained to reconstruct the input corrupted with noise much like a normal stacked autoencoder, however, the encoder and decoders themselves have three or four shallow layers. As opposed to greedy pretraining of the stacked denoising autoencoder, in a deep autoencoder the weights for all layers in the network are trained jointly through backpropagation. Deep autoencoders were investigated in~\citet{feng2014speech} for noisy reverberant speech recognition, and ~\citet{shao2015sparse} for missing modality face recognition. They are known to often yield more interesting features than greedily trained stacked autoencoders, which motivates us to explore deep autoencoders along with stacked architectures for learning affective representations.

\subsection{Recurrent Autoencoder}

A recurrent neural network~\citep{elman1991distributed} is widely used for learning from temporal data, where the $k$-th hidden layer of the network at time $t$ is a function of the $k-1$-th hidden layer at time $t$, and the $k$-th layer itself at the previous time step $t-1$, as shown below for a multi-layer network:
\[
\mathbf{h}_{t}^{k} = f(\mathbf{h}_{t}^{k-1},\mathbf{h}_{t-1}^{k})
\]
 The output of the network at time $t$, $\mathbf{y}_{t}$ is a function of the hidden layer below it. When the recurrent neural network is trained as an autoencoder, the target sequence is set to be equal to the input sequence.  An RNN can be trained using backpropagation through time, however the training suffers from the vanishing gradient problem. To address this issue, ~\citet{hochreiter1997long} proposed LSTM (Long Short Term Memory), which can model long-term temporal dependencies, and do not suffer from vanishing gradients. In our experiments, we use single-hidden layer recurrent BLSTM (Bidirectional LSTM) autoencoders to learn dimensionally reduced representations at utterance level, where the representation can be obtained from the BLSTM cell activation for the last frame of the utterance, or by averaging the hidden layer BLSTM cell activations across all frames. We have chosen to generate utterance level representations through an average over all frames, as visualization experiments indicate that this leads to better discrimination between emotion categories. Similar to the denoising autoencoder, the SSE(Sum of Squared Error Loss) is used for training the recurrent autoencoder. 

\subsection{Dataset}

For our experiments, we used the IEMOCAP dataset ~\citep{busso2008iemocap}, which is a well-known dataset for speech emotion recognition comprising of acted and spontaneous multimodal interactions of dyadic sessions between actors, where conversations are scripted to elicit emotional expressions. The dataset consists of around 12 hours of speech from 10 human subjects, and is labeled by three annotators for emotions such as \textit{Happy}, \textit{Sad}, \textit{Angry}, \textit{Excitement}, \textit{Neutral} and \textit{Surprise}, along with dimensional labels such as \textit{Valence} and \textit{Activation}. We use all the utterances for unsupervised pre-training of the autoencoder but perform classification experiments only on four emotions - Neutral, Angry, Sad and Happy. Each utterance is approximately 2-5 seconds in duration,  with short periods of silence before and after speech in all utterances. Prior to the classification experiments, for purposes of visualization we split the dataset into seven subjects (training set), and three subjects (validation set). Each utterance is annotated by three annotators for emotion and affective dimensions. For the supervised classification task, we have considered only utterances which have at least two annotators in agreement about the utterance emotion. The affective dimensions are annotated on a Likert scale of 1-5, where we have averaged out the dimension ratings across all three annotators. For unsupervised pre-training of the stacked denoising autoencoder, we have trained over the entire dataset of around 10,000 utterances. 

\subsection{Features Extracted}

We extracted spectrograms from the utterances, with a frame width of 20 ms and a frame overlap of 10 ms. In two different settings, we extract 513 and 128 FFT (Fast Fourier Transform) bins, since we wish to investigate whether fine-grained frequency information is necessary to produce better affective representations. We used a log-scale in the frequency domain, since a higher emphasis in lower frequencies has been shown to be more significant for auditory perception. Mel filterbanks have also been successfully used for speech recognition~\citep{sainath2013learning}, thus log-Mel spectrograms~\citep{ezzat2008discriminative} are also extracted from the utterances, with 40 filters in the Mel filterbank. Contrary to other works in the domain of speech emotion recognition, we do not directly extract any off-the-shelf features, such as Mel-Frequency Cepstral Coefficients (MFCCs), or voice quality features ~\citep{scherer2013investigating}, since we wish to investigate if the autoencoder is useful at discovering affective features directly from the frequency representation of the speech signal. However, we use off-the-shelf features from the COVAREP toolbox~\citep{degottex2014covarep} for comparison with the learned representations in an emotion recognition task.  

\subsection{Temporal Context Windows}

Each frame of the spectrogram is 20 ms in duration, which is generally accepted in the speech recognition community~\citep{huang2001spoken}. However, for learning emotions and other affective attributes, prior work has shown that longer temporal windows of the order of a hundreds of milliseconds is useful~\citep{kim2013emotion},~\citep{han2014speech}. Consequently, we explore the concept of temporal context windows, where each input feature to the autoencoder is a window of consecutive spectral frames. If $K$ is a parameter denoting the length of past/future context, and $\mathbf{X}_t$ is a spectrogram frame at time $t$, the input temporal context window at time $t$ is $[\mathbf{X}_{t-K},...,\mathbf{X}_{t-2},\mathbf{X}_{t-1},\mathbf{X}_t,\mathbf{X}_{t+1},\mathbf{X}_{t+2},...,\mathbf{X}_{t+K}]$. In this paper, we have conducted experiments with $K=2$, which would correspond to a context window size of $2K+1 = 5$. We have not performed any experiments with temporal context windows for 513 FFT bins, since the high dimensionality of the resulting windows might cause overfitting when training the autoencoder models.

\begin{table}[t]
\caption{Utterance level classification performance for different spectrogram and autoencoder architecture combinations on a held-out validation set (best performing models selected for visualization are indicated in bold) }
\label{config-autoencoder}
\begin{center}
\begin{tabular}{lllll}
\multicolumn{1}{c}{\bf Spectrogram}  &\multicolumn{1}{c}{\bf Autoencoder } &\multicolumn{1}{c}{\bf Context} &\multicolumn{1}{c}{\bf Architecture} &\multicolumn{1}{c}{\bf Accuracy}  \\
\hline 
  & \textbf{Tied} & \textbf{1} &  \textbf{513-256-128-64-4}  & \textbf{46.23\%} \\
513-bin FFT  & Untied & 1 & 513-256-128-64-4   & 45.76\%  \\
  & Deep & 1 & 513-256-128-64-4  & 46.35\%  \\
\hline
   &  Tied   &  1 & 128-64-32-4  & 44.14\%  \\
   &  Untied   & 1       &   128-64-32-4     &  43.45\%       \\
   &  Deep   &   1     &   128-64-32-4     &    43.25\%     \\
128-bin FFT    &  \textbf{Tied}  &   \textbf{5}     &  \textbf{640-320-160-80-4}      & \textbf{50.39\%}  \\
    &  Untied   &  5      &  640-320-160-80-4      &49.081\%         \\
    &  Deep   &   5     &    640-320-160-80-4    &   48.76\%       \\
\hline
 & Softmax   &   1   &   40-4  &      44.45\%     \\
Log-Mel &  Tied  &   5   &  200-100-50-4   &  43.69\%          \\
 & Untied   &   5   &  200-100-50-4   &     43.76\%      \\
  & \textbf{Deep}   &  \textbf{5}    &  \textbf{200-100-50-4}   &   \textbf{44.07\%}        \\
\hline 
\end{tabular}
\end{center}
\end{table}

\section{Validation Experiments to compare Model Configurations}

In the previous section, we described different spectrogram features for input to the autoencoder, as well as various autoencoder architectures (such as stacked tied/untied weights and deep autoencoders). In this section, we describe our approach to find the best features and autoencoder models through empirical classification experiments. We split the IEMOCAP dataset into a training set of 7 speakers, and a held out validation set of 3 speakers. Table~\ref{config-autoencoder} lists the configurations we have used for autoencoder pre-training, followed by a classification stage where a simple softmax classifier is trained on the extracted bottleneck features till convergence. Similar to most prior work in this domain~\citep{masci2013multi}, we used a pyramidal stacking approach for the autoencoders, where the number of neurons is halved for the next higher layer. We also report the held-out validation accuracies, and select three configurations for the visualization experiments, which we describe in Section 5. For all architectures, the autoencoder is pretrained with batched stochastic gradient descent for 5 epochs, with a learning rate of 1e-4, a weight decay of 1e-4, and batch size of 500. 20\% of the features in the frame are randomly dropped when training the denoising autoencoders.  
From an examination of Table~\ref{config-autoencoder}, we find that the best classification accuracy is obtained using 128-bin FFT spectrograms, with a temporal context window of 5 frames, and a stacked denoising architecture with tied weights. The accuracies obtained with log-Mel spectrograms are lower than those obtained with FFT spectrograms, which indicate that fine detail in the spectrum might be necessary for emotion classification. We also observe that while deep autoencoder features are better than those obtained from the stacked architectures, the improvement in performance is marginal. Tying weights also seem to have little or no improvement in performance.

\section{Visualization of unsupervised representations}

In this section, we describe the visualization experiments we conduct to obtain an insight into the ability of the autoencoder architectures to learn features which are discriminative of affective attributes. The activations learnt at the bottleneck layer of the autoencoders are dimensionally reduced using tSNE ~\citep{van2008visualizing} and are visualized at frame and utterance level, with the primary emotion categories. We also examine the variation of the learnt representations with two affective dimensions : (1) Activation, which is a measure of intensity of the emotion, and (2) Valence, which is a measure of sentiment of the utterance. We choose the following spectrogram-autoencoder configurations for the visualization experiments: \\
(1) \textit{DEEP-MEL-5} : Deep autoencoder trained on log-mel spectrograms with context window length of 5. \\
(2) \textit{TIED-128-5} : Stacked autoencoder (tied weights) trained on 128 bin FFT spectrograms with context window length of 5. \\
(3) \textit{TIED-513-1} : Stacked autoencoder (tied weights) trained on 513 bin FFT spectrograms (single frames with no context windows). \\
\subsection{Frame based representations with the denoising autoencoder}
\begin{figure}[th]
  \subfloat[Emotion : \textit{DEEP-MEL-5}]{\includegraphics[scale=0.25]{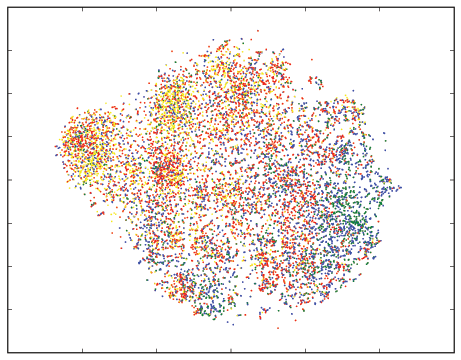}\label{fig:f1_rnn}}
	\hfill
	\subfloat[Emotion : \textit{TIED-128-5}]{\includegraphics[scale=0.25]{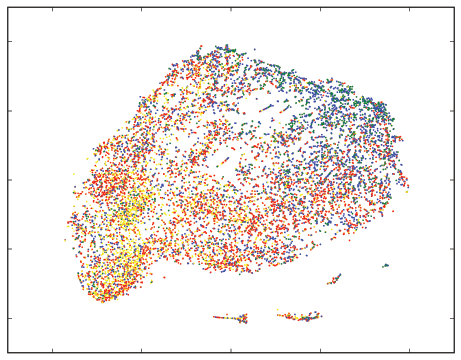}\label{fig:f4_rnn}}
  \hfill
	\subfloat[Emotion : \textit{TIED-513-1}]{\includegraphics[scale=0.25]{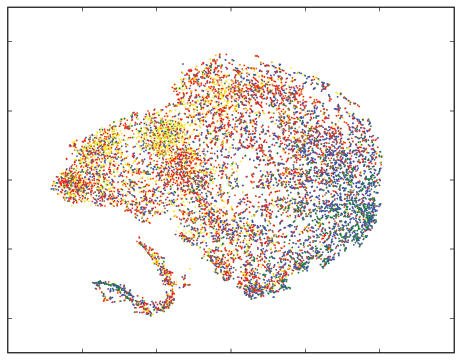}\label{fig:f7_rnn}} \\
  \hfill
	\subfloat[Activation : \textit{DEEP-MEL-5}]{\includegraphics[scale=0.25]{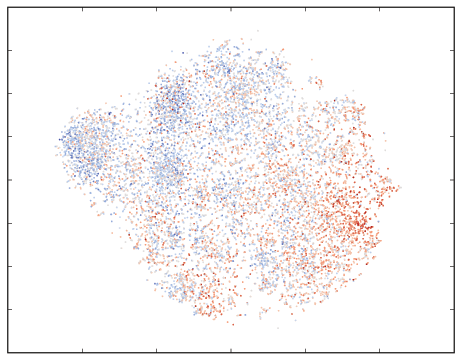}\label{fig:f3_rnn}} 
  \hfill
	\subfloat[Activation : \textit{TIED-128-5}]{\includegraphics[scale=0.25]{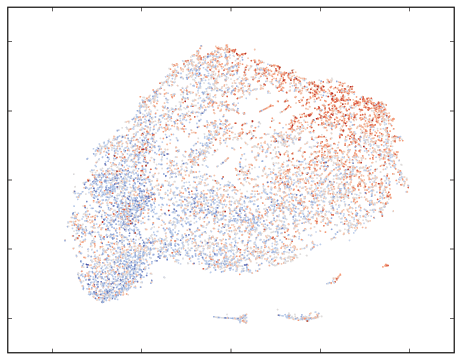}\label{fig:f6_rnn}} 
	\hfill
	\subfloat[Activation : \textit{TIED-513-1}]{\includegraphics[scale=0.25]{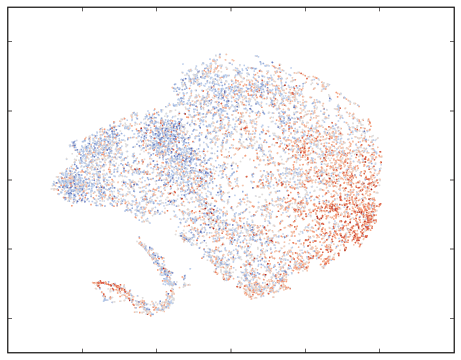}\label{fig:f9_rnn}} \\
	\hfill
	\subfloat[Valence : \textit{DEEP-MEL-5}]{\includegraphics[scale=0.25]{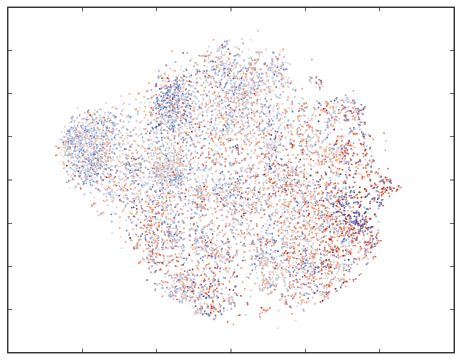}\label{fig:f13_rnn}} 
  \hfill
	\subfloat[Valence :  \textit{TIED-128-5}]{\includegraphics[scale=0.25]{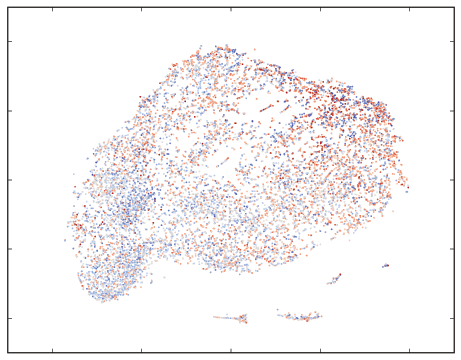}\label{fig:f16_rnn}} 
	\hfill
	\subfloat[Valence : \textit{TIED-513-1}]{\includegraphics[scale=0.25]{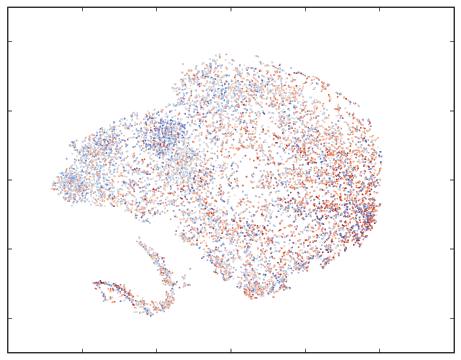}\label{fig:f19_rnn}}
  \caption{Visualization of frame-level representations of emotion, activation and valence learned by model configurations \textit{DEEP-MEL-5}, \textit{TIED-128-5} and \textit{TIED-513-1}. The intensity gradient for each dimension is from blue (lowest) to red (highest). Yellow - Sadness; Red - Neutral; Blue - Happiness; Green - Angry}
	\label{fig:scat_auto}
\end{figure}

Figure~\ref{fig:scat_auto} shows scatter plots of the frame-level representations for four primary emotions - (1) Happy (2) Sad (3) Angry and (4) Neutral. The visualization is at a frame level, which gives us an insight into the acoustic characteristics of the speech, as well as the discriminative ability of the autoencoder to learn affective characteristics from the data. All frames within an utterance are assigned the emotion label of the entire utterance. While this is not entirely true in practice, we have observed that it is capable of providing meaningful visualizations. On an initial examination of the scatter plots, we find that smaller islands (regions disconnected from the biggest cluster) occur where the spectral characteristics of the frames are markedly different from the rest of the utterance. This mostly corresponds to the separation between the silence/pauses and speech within the utterances. Further the unsupervised representations can clearly separate out anger (in green) from sadness (in yellow), which gets grouped into tight clusters. Happiness (in blue) and anger (in green) have low-variance clusters, while the high variance neutral category does not have a well-defined cluster. This is intuitive, since from prior psychological studies~\citep{scherer2005emotions}, the neutral emotion is not well-defined in terms of spectral characteristics, and also depends largely on speaker specific attributes. \\

We also explore whether the autoencoder can learn affective attributes from the speech such as activation (intensity of emotions) and valence (positive or negative sentiments). Figure \ref{fig:scat_auto} also shows scatter plot of learned representations, colored for each attribute according to intensity (blue for lowest and red for highest). We find that the autoencoder is most discriminative of activation, followed by valence. The sensitivity of the representation to activation explains why anger and sadness are clustered and well separated. On examining the scatterplots for valence, we find that low valence utterances (sadness and anger) are more towards the edges, compared to medium and high valence utterances.

\subsection{Utterance based representations with the recurrent autoencoder}
\begin{figure}[th]

  \subfloat[Emotion : \textit{DEEP-MEL-5}]{\includegraphics[scale=0.25]{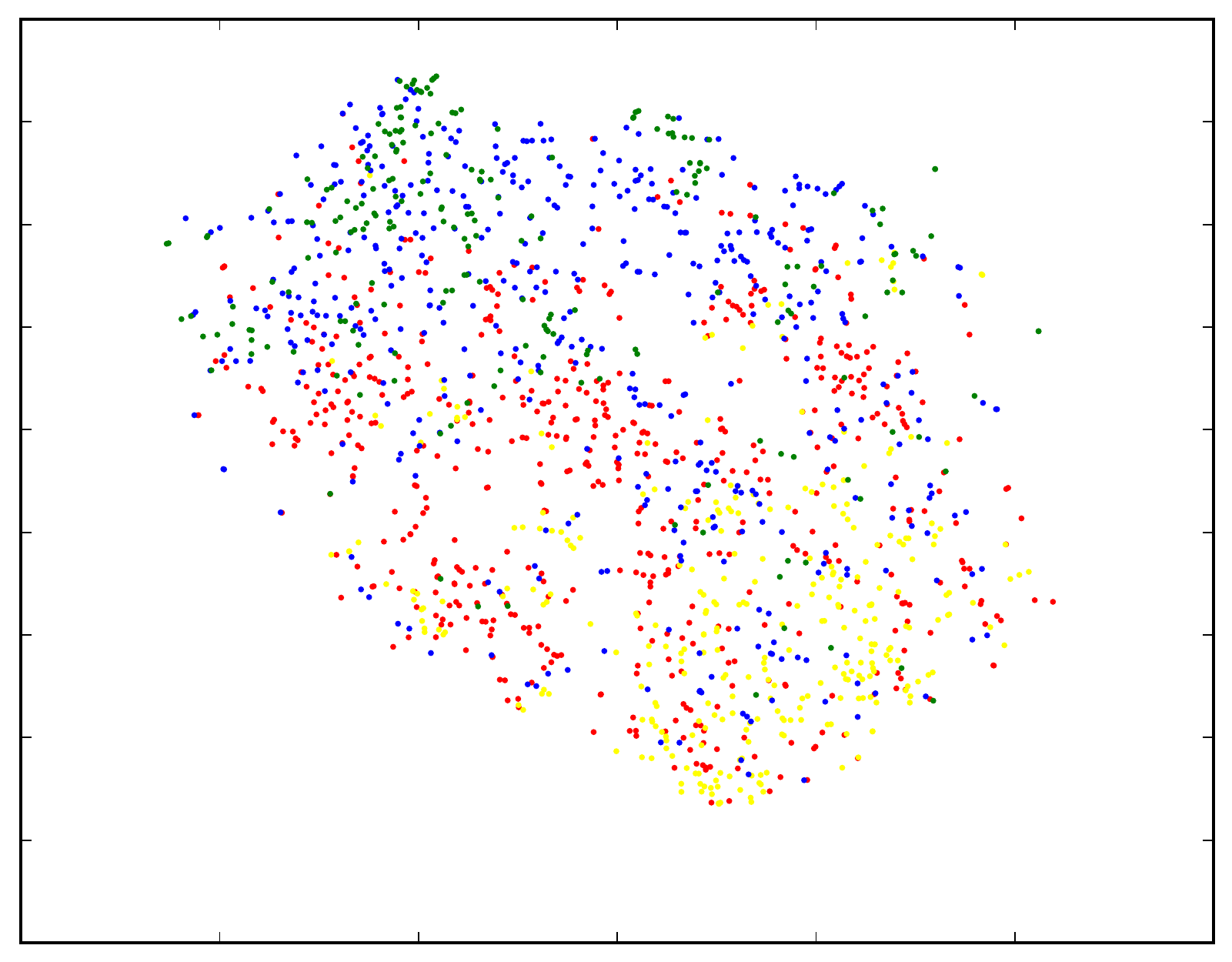}\label{fig:f1_rnn}}
	\hfill
	\subfloat[Emotion : \textit{TIED-128-5}]{\includegraphics[scale=0.25]{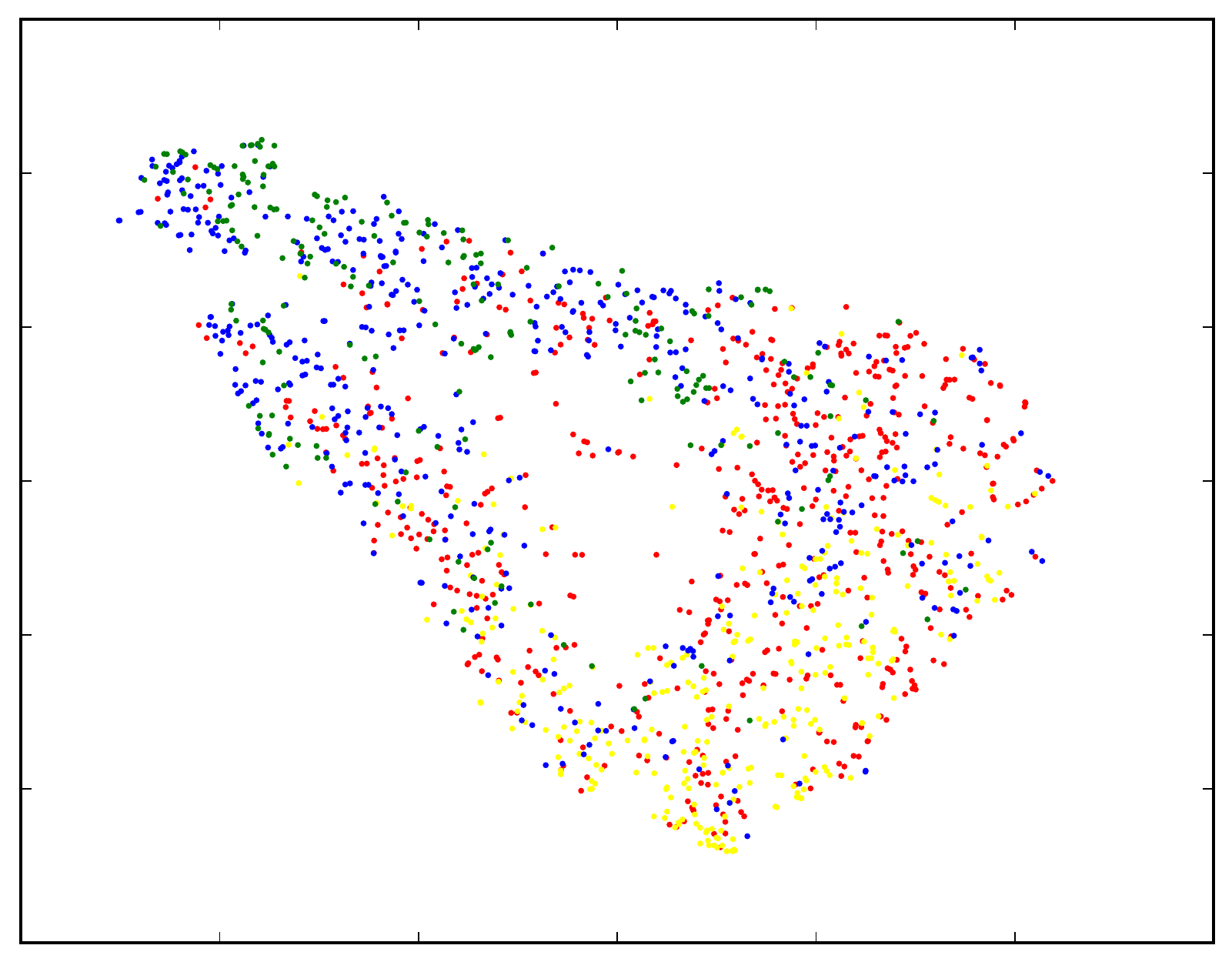}\label{fig:f4_rnn}}
  \hfill
	\subfloat[Emotion : \textit{TIED-513-1}]{\includegraphics[scale=0.25]{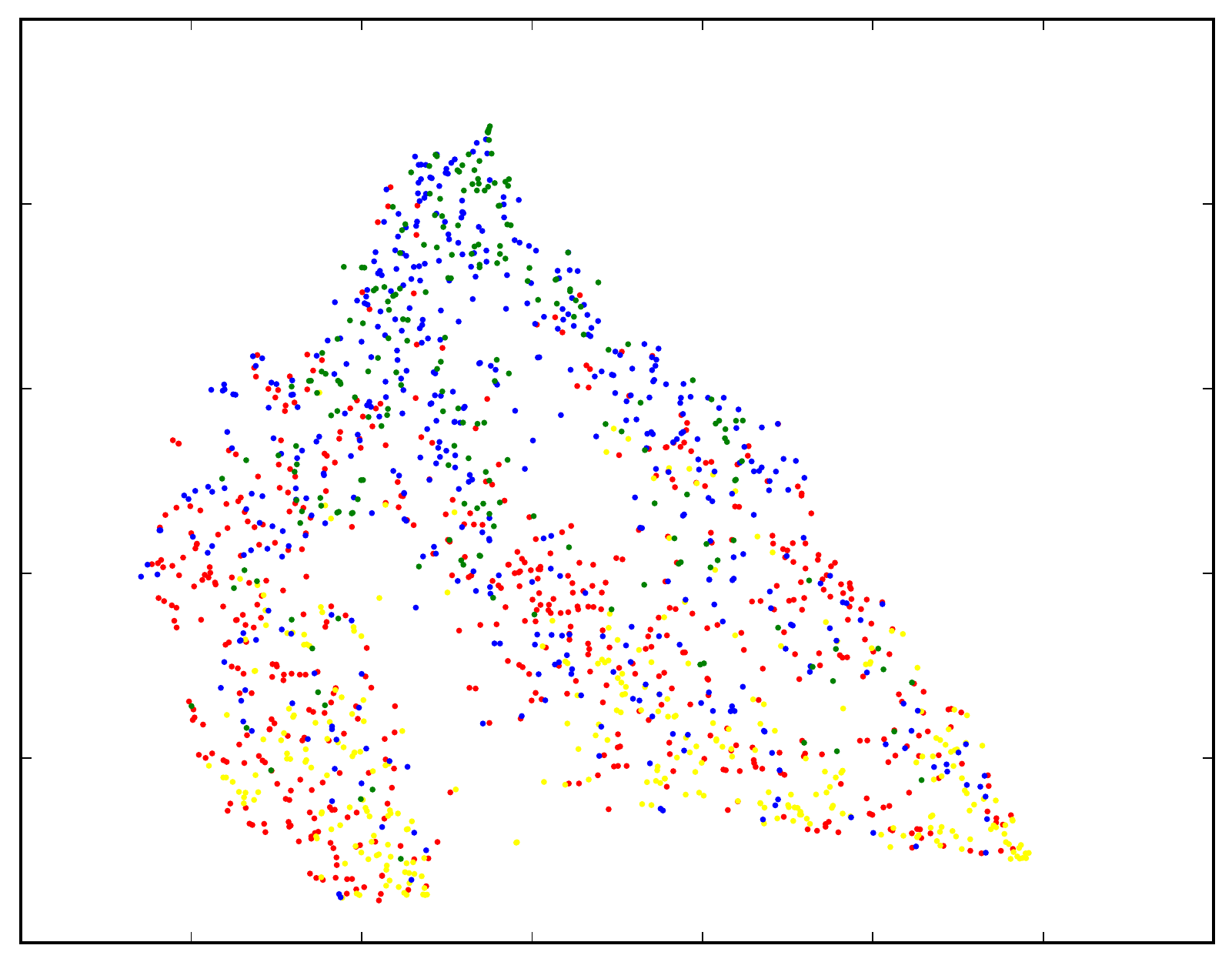}\label{fig:f7_rnn}} 
  \\
	\subfloat[Activation : \textit{DEEP-MEL-5}]{\includegraphics[scale=0.25]{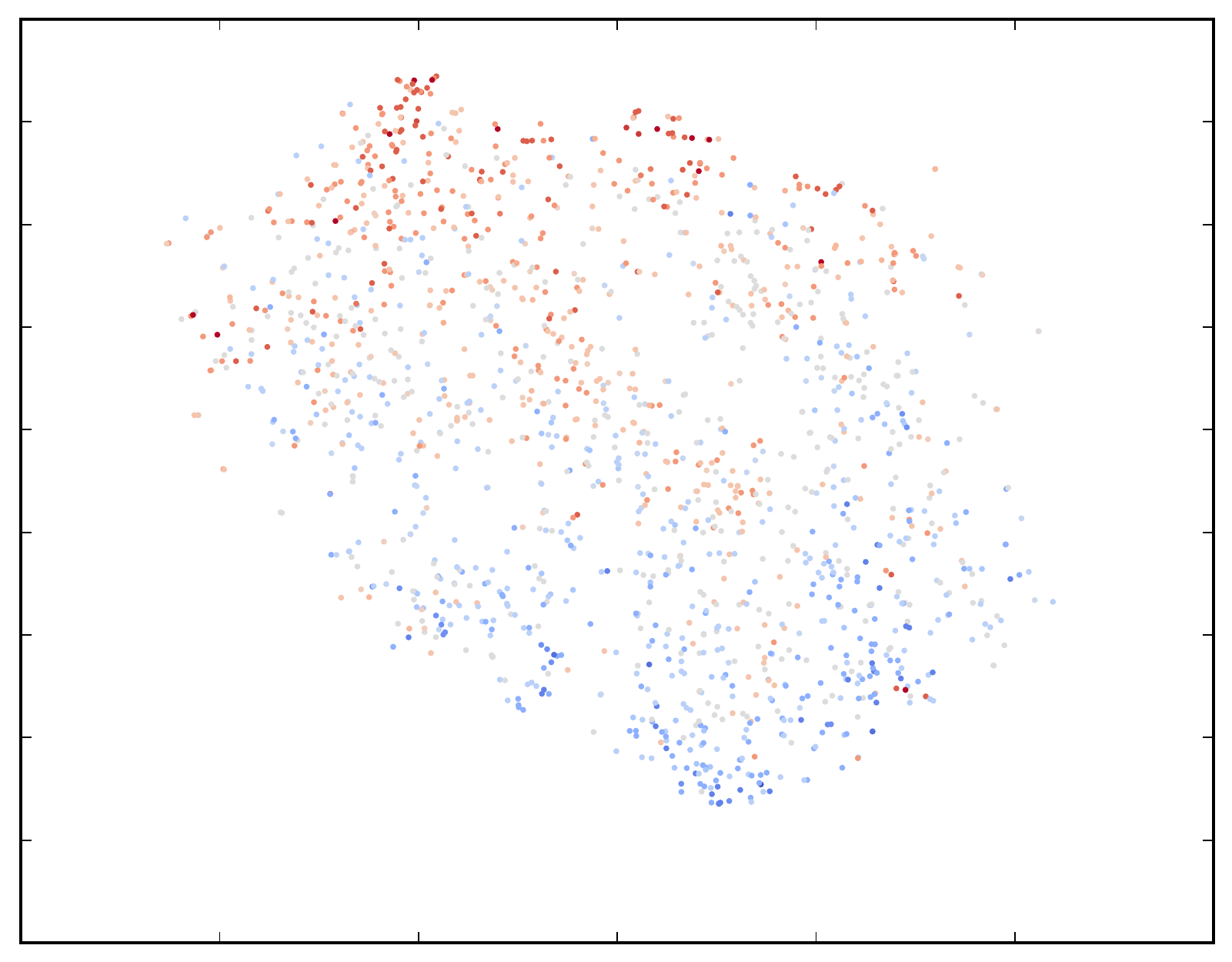}\label{fig:f3_rnn}} 
  \hfill
	\subfloat[Activation : \textit{TIED-128-5}]{\includegraphics[scale=0.25]{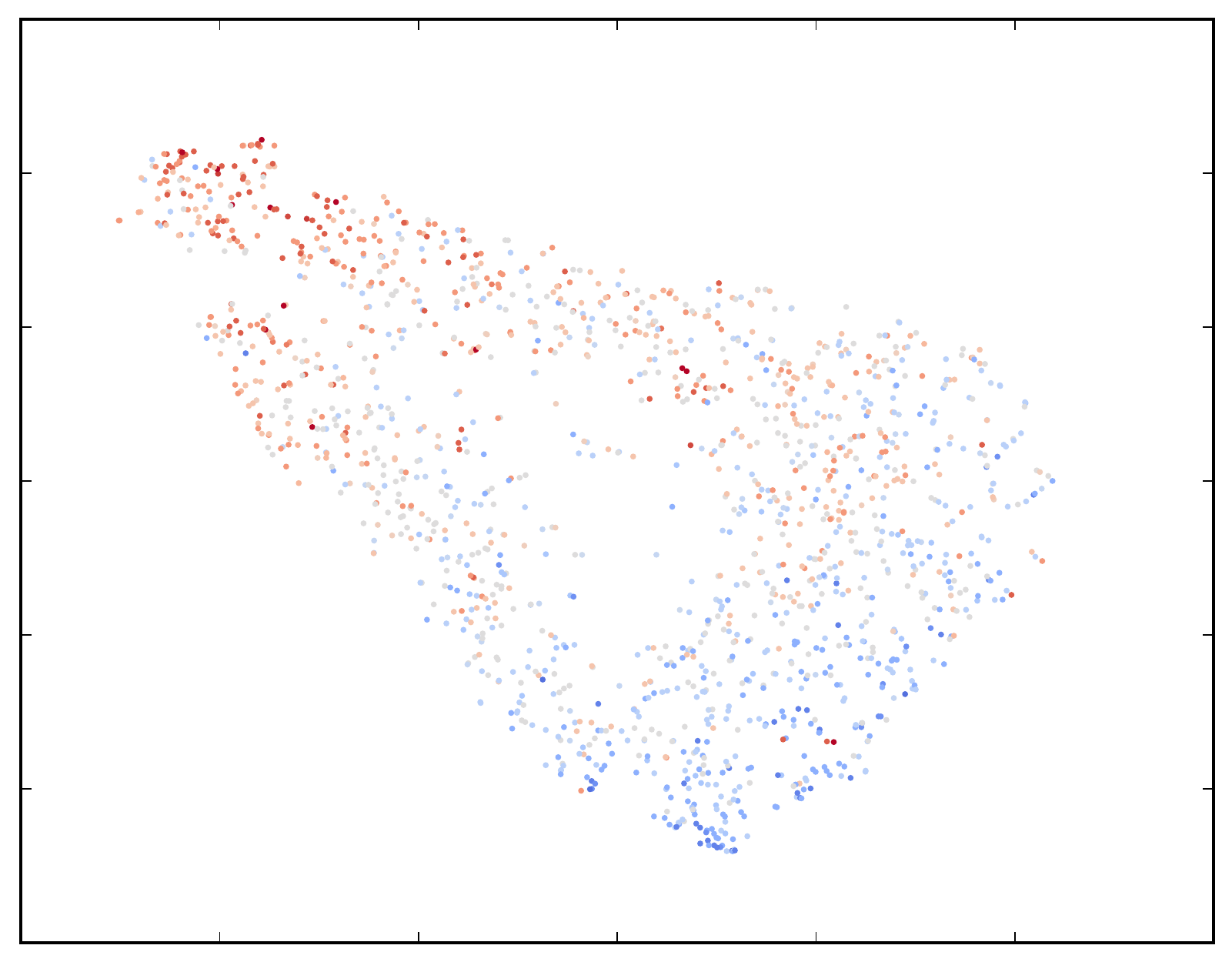}\label{fig:f6_rnn}} 
	\hfill
	\subfloat[Activation: \textit{TIED-513-1}]{\includegraphics[scale=0.25]{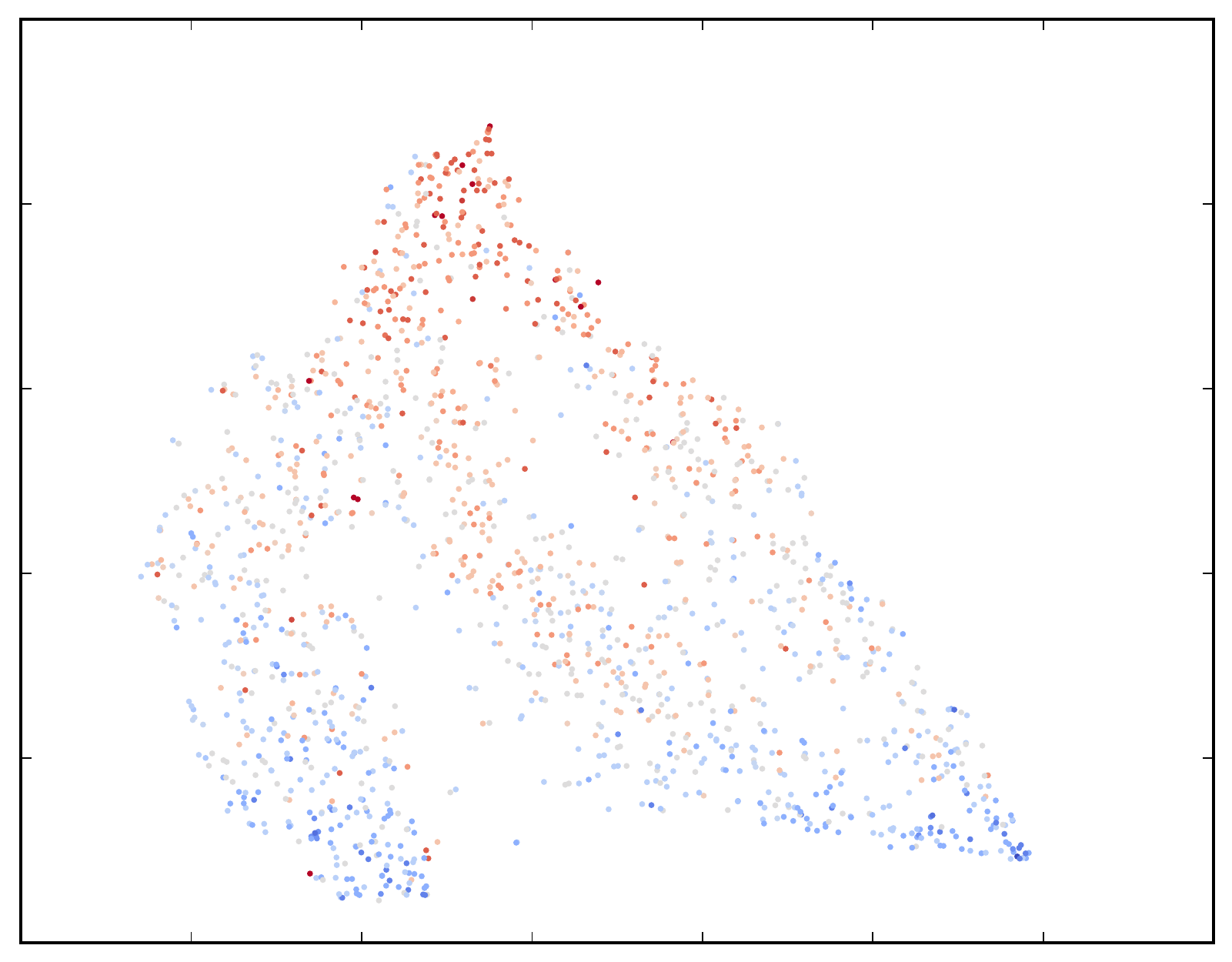}\label{fig:f9_rnn}} 
	\\
	\subfloat[Valence : \textit{DEEP-MEL-5}]{\includegraphics[scale=0.25]{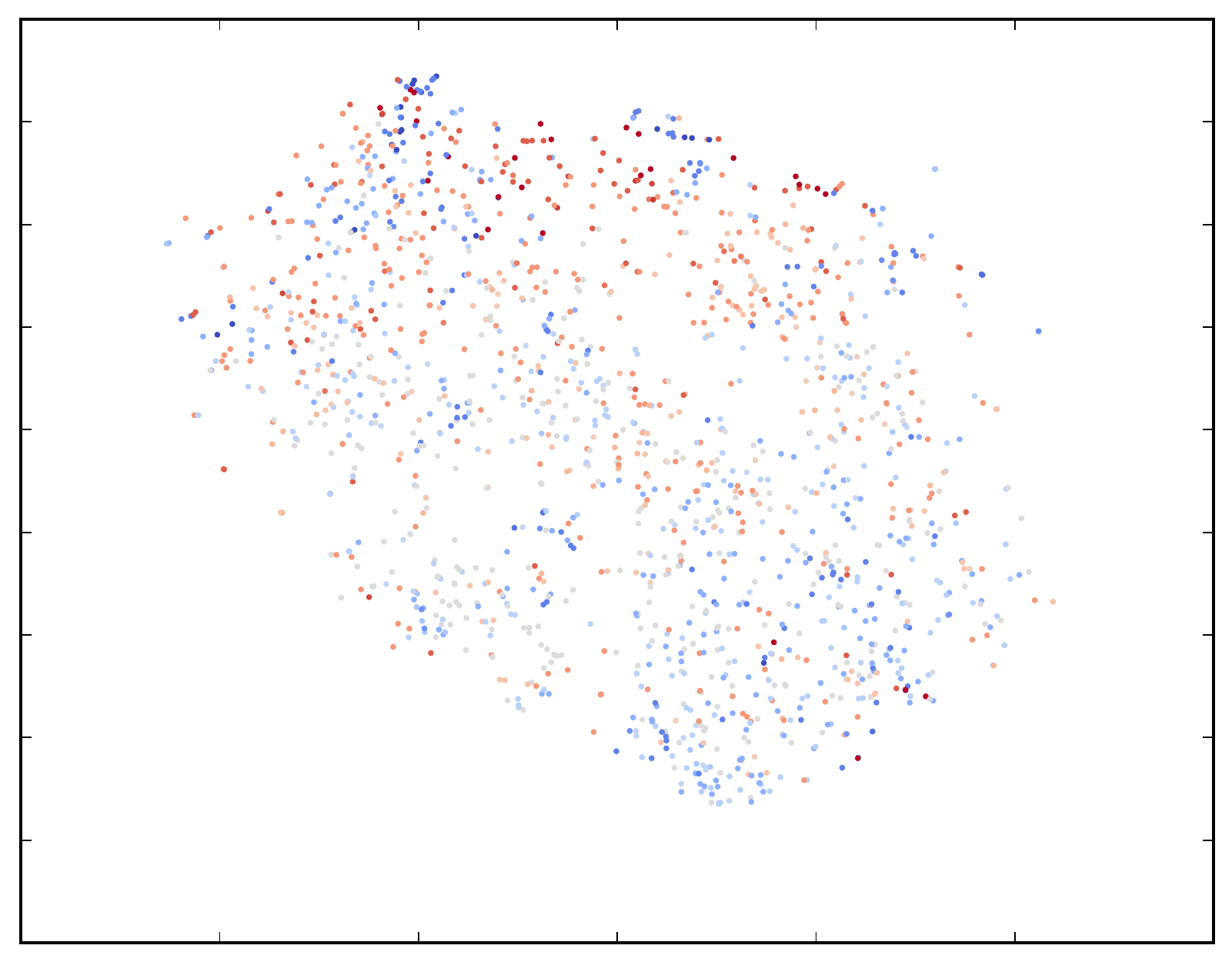}\label{fig:f13_rnn}} 
  \hfill
	\subfloat[Valence :  \textit{TIED-128-5}]{\includegraphics[scale=0.25]{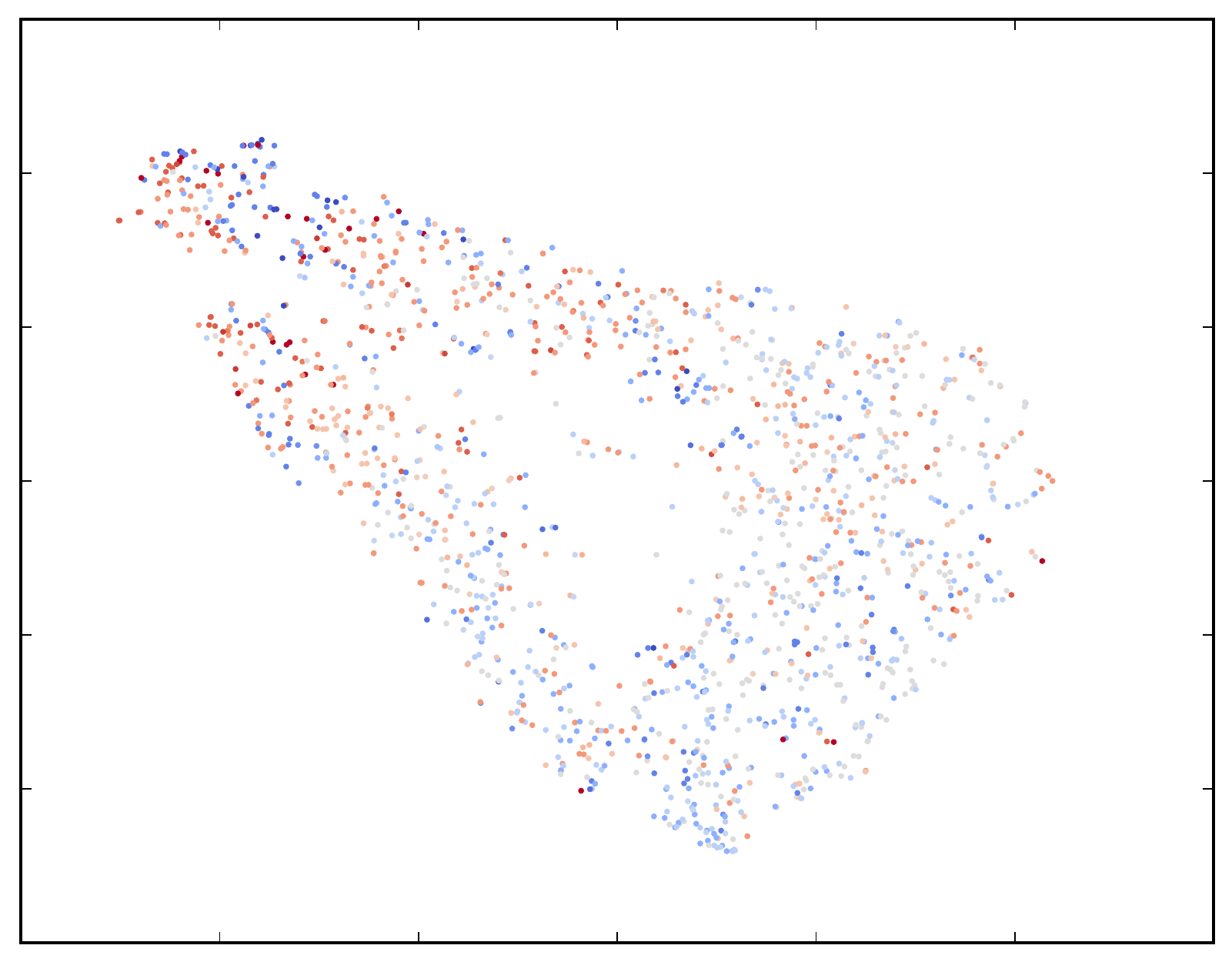}\label{fig:f16_rnn}} 
	\hfill
	\subfloat[Valence : \textit{TIED-513-1}]{\includegraphics[scale=0.25]{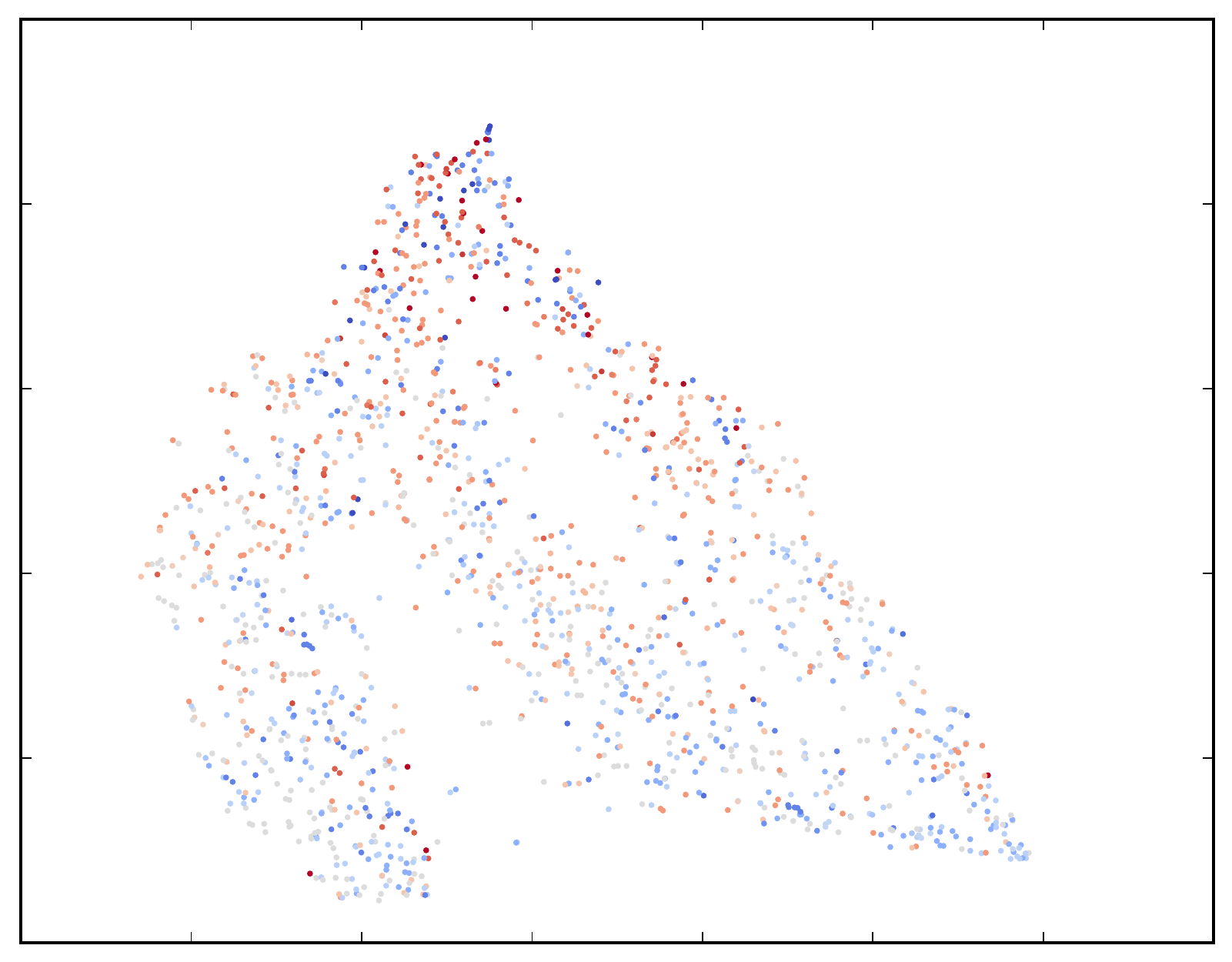}\label{fig:f19_rnn}}
  \caption{Visualization of recurrent autoencoder learnt utterance-level representations of emotion, activation and valence for model configurations \textit{DEEP-MEL-5}, \textit{TIED-128-5} and \textit{TIED-513-1}. The intensity gradient for each dimension is from blue (lowest) to red (highest). Yellow - Sadness; Red - Neutral; Blue - Happiness; Green - Angry}
	\label{fig:scat_rnn}
\end{figure}

We have visualized scatter plots of categorical emotions, and affective dimensions such as activation and valence at frame level. However, when the entire utterance is annotated with a single attribute, such as emotion or valence/activation, it would be necessary to learn a representation for the entire utterance. This would not only filter out phonetic information which varies across frames in an utterance, but also be convenient for visualization. We have used a recurrent neural network with a single hidden BLSTM (Bidirectional Long Short Term Memory) layer as an autoencoder to learn a temporal hidden representation which can map the utterance to itself. The BLSTM-RNN is added on top of the denoising autoencoder, i.e. it is trained with a learning rate of 1e-6 over 45 epochs on the bottleneck-layer activations obtained by the autoencoder. 16-dimensional frame-level hidden BLSTM cell activations are extracted from the trained recurrent autoencoder. Our experiments indicate that averaging the framewise activations to generate an utterance-level representation produces representations visually more indicative of affect, rather than selecting the cell activations at the last frame. Figure~\ref{fig:scat_rnn} shows utterance-level scatter plots of emotions and other affective attributes such as valence and activation. From the plots we find a similar correlation with affective traits as in the frame based plots, indicating that the recurrent autoencoder is effective at learning utterance representations.

\section{Emotion Classification with Learnt Representations}

The features extracted from the autoencoder are used for emotion classification on the IEMOCAP dataset. In the previous section, we have described our work in unsupervised pretraining of denoising and recurrent autoencoders with different spectrograms and architectures. We now use the pretrained layers of the best performing configuration on the validation experiments (\textit{TIED-128-5}) to initialize weights of a MLP (Multi-layer Perceptron). To evaluate the performance of the learned representations, we add a softmax layer with four output units, for each of the four main emotions (\textit{Angry},\textit{Happy},\textit{Neutral} and \textit{Sadness}). Let us represent the dataset by a collection of speech frames $\{\mathbf{x}_{ij}\}$ where $i$ denotes the utterance index, and $j$, the time frame in the $i$-th utterance. If $\mathbf{y}_{ij}=f(\mathbf{x}_{ij})$ is the activation of the top layer in $C=4$ emotion category dimensions predicted by the MLP, then the predicted emotion category $C_i$ for the $i$-th utterance is given by:
\[
C_i = \mathbf{argmax}_{k\in 1:C} \sum_{j=1}^{j=T_i} \mathbf{y}_{ij}(k)
\]
Similar to the approach in ~\citet{han2014speech} we split the dataset (comprising of five sessions with two distinct speakers in each session) into five folds where for each fold, data from four sessions  is used as the training set, and the remaining session for validation and testing.  For the session we choose for validation and testing, we consider one speaker for validation of hyper-parameters, and the remaining speaker for testing. We average the accuracy across all five folds, and present both weighted and unweighted accuracies in Table~\ref{accuracies}. We compare our results with three approaches:
(1) A network with the same architecture without pre-training (2)
A softmax classifier trained on features extracted from the COVAREP toolbox, such as MFCCs and prosodic features (for example, pitch, peak slope, Normalized Amplitude Quotient (NAQ), and difference between first two harmonics in speech (H1-H2)) and (3) the DNN-ELM approach described in ~\citet{han2014speech}, for which results in the same experimental setting (emotion categories and speaker-independent data splits) as our work have been reported in ~\citet{lee2015high}. The \textit{Angry} and \textit{Neutral} categories exhibit much lower performance than the \textit{Happy} and \textit{Sad} categories, which can be understood from the presented scatter plots in Figure~\ref{fig:scat_rnn}, as the \textit{Angry} class has an overlap with \textit{Happy} and \textit{Neutral} categories, and \textit{Neutral} has an overlap with all the other categories. \citep{lee2015high} report an unweighted and weighted accuracy of 52.13\%, and 57.91\% respectively for the DNN-ELM model. It is worth noting that our work is more focused on finding a good unsupervised representation for affect, rather than utilizing a highly complex model for subsequent classification.  

\label{others}

\begin{table}[t]
\caption{Emotion classification accuracy (utterance level) for our approach, compared with  (1) MLP without pre-training (2) softmax classifier trained on COVAREP features \iffalse and DNN-ELM \fi}
\label{accuracies}
\begin{center}
\begin{tabular}{lllll}
\multicolumn{1}{c}{\bf Emotion} &\multicolumn{1}{c}{\bf \textit{TIED-128-5}} &\multicolumn{1}{c}{\bf \textit{MLP}} &\multicolumn{1}{c}{\bf \textit{COVAREP}} \\
\hline \\
Happy         & 36.04 & 41.56 & 31.03 \\
Angry             & 44.95 & 24.03 &  32.31     \\
Sad             & 73.80  & 66.55 & 50.17  \\
Neutral         &  41.59 & 48.58 & 61.48  \\ 
\hline 
Weighted Accuracy     & 48.10 & 46.75  & 44.67   \\
Unweighted Accuracy   &  49.09 & 45.18  & 43.74  \\
\end{tabular}
\end{center}
\end{table}

\section{Future Work and Conclusions}
In this paper, we explore stacked and deep denoising autoencoders for representation learning of affective and paralinguistic attributes from speech. We have also experimented with spectrogram types and temporal context windows, and found that a stacked tied autoencoder with 128 FFT bins and a temporal context window of 5 performs the best on a speaker-independent held-out validation set. We find that the representations learnt by the autoencoder are sensitive to the four primary emotions, and to activation as an affective dimension. Emotion classification experiments show that a multi-layer perceptron, along with layers initialized through pre-training achieve an average weighted classification accuracy of around 48.10\% on a five-fold leave-one-session-out testing scheme. We also train BLSTM-autoencoders to visualize utterance specific representations. For future work, we wish to explore BLSTM classifiers, and inverse filtering for speech emotion classification.

\subsubsection*{Acknowledgments}
The work depicted here is sponsored by the U.S. Army Research Laboratory (ARL) under contract number W911NF-14-D-0005 and the National Science Foundation under Grant No. IIS-1118018. Statements and opinions expressed and content included do not necessarily reflect the position or the policy of the National Science Foundation or the Government and no official endorsement should be inferred. Sayan Ghosh also acknowledges the USC Viterbi Graduate School Fellowship for funding his graduate studies. The Tesla K40 used for this research was donated by the NVIDIA Corporation.

\bibliography{iclr2016_conference}
\bibliographystyle{iclr2016_conference}

\end{document}